\documentclass{article} 
\usepackage{arxiv_preprint, times}

\usepackage{amsmath,amsfonts,bm}

\def\eqref#1{equation~\ref{#1}}

\def\1{\bm{1}}

\DeclareMathAlphabet{\mathsfit}{\encodingdefault}{\sfdefault}{m}{sl}
\SetMathAlphabet{\mathsfit}{bold}{\encodingdefault}{\sfdefault}{bx}{n}

\usepackage{hyperref}
\usepackage{url}
\usepackage{graphicx}
\usepackage{colortbl}
\usepackage{booktabs} 
\usepackage{multirow}
\usepackage{subfigure}
\usepackage{fontawesome}

\def\mathbi#1{\textbf{\em #1}}

\newcommand{\txt}[1]{{\texttt{#1}}}

\definecolor{grey}{RGB}{128,138,135}
\definecolor{darkgrey}{RGB}{96,96,96}
\definecolor{lavender}{HTML}{E5E2FB}
\definecolor{lightblue}{HTML}{CEEBF9}
\definecolor{lightyellow}{HTML}{F7D9AE}
\definecolor{lightred}{HTML}{EEDCDB}
\definecolor{green}{HTML}{3BCB41}
\definecolor{darkgreen}{HTML}{156C09}
\definecolor{purple}{HTML}{9903F0}

\title{CLIP model is an Efficient Continual Learner}
\author{
    Vishal Thengane\textsuperscript{1, \faEnvelopeO}, 
        Salman Khan\textsuperscript{1,2}, 
        Munawar Hayat\textsuperscript{3}, 
        Fahad Khan\textsuperscript{1,4} \\
    \textsuperscript{1}Mohamed bin Zayed University of Artificial Intelligence, UAE \\
    \textsuperscript{2}Australian National University, Australia \\
    \textsuperscript{3}Monash University, Australia \\
    \textsuperscript{4}Linköping University, Sweden \\
    \textsuperscript{\faEnvelopeO}\href
        {mailto:vishal.thengane@mbzuai.ac.ae}{\texttt{vishal.thengane@mbzuai.ac.ae}
    } 
    \and
}

\begin{document}

\maketitle

\begin{abstract}

The continual learning setting aims to learn new tasks over time without forgetting the previous ones. The literature reports several significant efforts to tackle this problem with limited or no access to previous task data. Among such efforts, typical solutions offer sophisticated techniques involving memory replay, knowledge distillation, model regularization, and dynamic network expansion. The resulting methods have a retraining cost at each learning task, dedicated memory requirements, and setting-specific design choices. In this work, we show that a frozen CLIP (Contrastive Language-Image Pretraining) model offers astounding continual learning performance without any fine-tuning (zero-shot evaluation). We evaluate CLIP under a variety of settings including class-incremental, domain-incremental and task-agnostic incremental learning on five popular benchmarks (ImageNet-100 \& 1K, CORe50, CIFAR-100, and TinyImageNet). Without any bells and whistles, the CLIP model outperforms the state-of-the-art continual learning approaches in majority of the settings. We show the effect on CLIP model's performance by varying text inputs with simple prompt templates. To the best of our knowledge, this is the first work to report the CLIP zero-shot performance in a continual setting. We advocate the use of this strong yet embarrassingly simple baseline for future comparisons in the continual learning tasks. Code is available at \href{https://github.com/vgthengane/Continual-CLIP}{\texttt{https://github.com/vgthengane/Continual-CLIP}}.  

\end{abstract}

\section{Introduction}

Traditionally, deep neural networks (DNNs) trained in a supervised manner on training sets comprising of all the classes of interest have shown excellent results. 
Such models can presumably learn all the relevant features from the dataset in a single training episode. However, in real world, all data samples may not be available at once. To cater for such scenarios, continual learning provides a promising paradigm, since it enables learning where the data distribution shifts over time. DNNs trained on such incremental stream of data, however, suffer from catastrophic forgetting since the previous task data can not be accessed in its entirety \citep{mccloskey1989catastrophic}.  

In the literature, four popular continual learning protocols exist; Task-incremental learning (TIL) use task-specific neural networks, where the task identity is assumed to be known at the inference. Class-incremental learning (CIL) settings add classes in a sequence with the task identity unknown at the inference. In Domain-incremental learning (DIL), the number of classes remain the same but the data domain evolves over time. Task-free (or task-agnostic) continual learning (TFCL) is a more general setting where there are no explicit task boundaries and data can appear freely in the continual learning phases \citep{de2021continual}. The major challenge faced by all these methods is to avoid forgetting previously learned knowledge while updating on new data.

Several specialized methods have been developed in continual learning literature to reduce catastrophic forgetting. Among such methods, typical solutions offer sophisticated techniques involving memory replay \citep{rebuffi2017icarl, shin2017continual, lopez2017gradient}, knowledge distillation \citep{hinton2015distilling, li2017learning}, model regularization \citep{kirkpatrick2017overcoming}, parameter isolation \citep{mallya2018packnet, fernando2017pathnet}, and dynamic network expansion \citep{Yan_2021_CVPR, Douillard_2022_CVPR, douillard2020podnet}. The resulting methods have a retraining cost at each learning task, need dedicated memory for storing exemplars or past models, and involve complex hyper-parameter tuning which limits their practical utility. Furthermore, the above continual learning protocols are generally addressed separately and the existing approaches involve setting-specific design choices making them non-transferable across different continual learning settings. 

\begin{figure}
  \includegraphics[width=\linewidth]{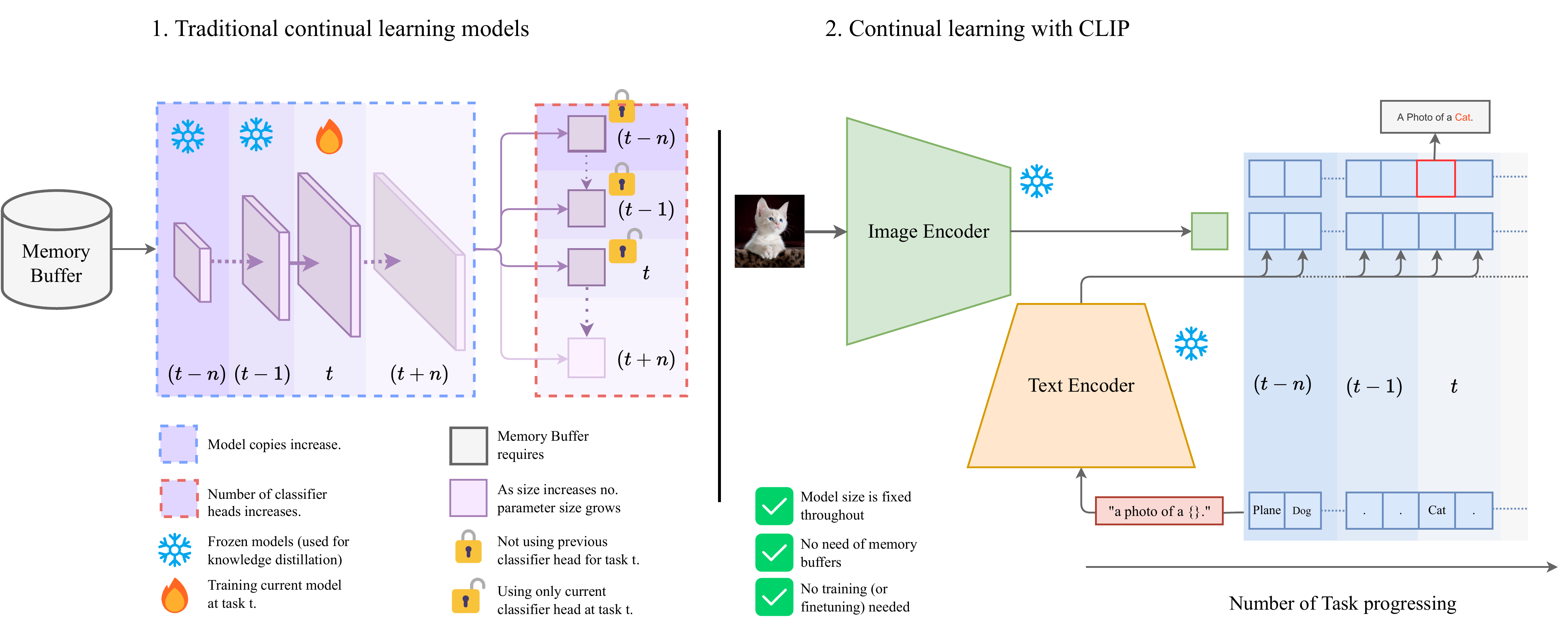}
  \caption{\emph{Left:} Traditional continual learning approaches can require memory buffers, complex hyper-parameter tuning, saving a copy of previous models and their number of classifier heads increases at each learning step. \emph{Right:} Our goal is to design a simplistic unified model that works well across multiple continual learning settings without incurring task-wise training, dedicated memory requirements and careful hyper-parameter selection. A CLIP-based continual model is shown to perform exceptionally well on a number of continual learning settings without requiring any training/fine-tuning, memory buffers or increase in model size with the growing number of tasks.}
  \label{fig:continual_clip}
\end{figure}

In this work, we aim to test the progress made so far towards a truly continual learning system. Our main question is to explore if the state-of-the-art narrow models can be replaced with a simple generic approach that does not require training for each incremental step, works without any exemplar memory storage and can work across all of the existing incremental learning protocols with minimal or no hyper-parameter tuning. To this end, we show that a frozen CLIP model \citep{radford2021learning} offers great promise due to its generalizable representations and zero-shot behaviour without requiring any parameter-tuning. We call the CLIP evaluated on a diverse set of continual learning settings as \textit{Continual-CLIP}. Figure \ref{fig:continual_clip} gives an overview of traditional continual learning methods and frozen CLIP in the continual learning system.

Our extensive evaluations across four diverse settings (TIL, CIL, DIL, and TFCL) and seven datasets (ImageNet-100 \& 1K \citep{deng2009imagenet}, CLEAR \citep{lin2021clear}, CIFAR100 \citep{krizhevsky2009learning}, TinyImageNet \citep{le2015tiny}, CoRe50 \citep{lomonaco2017core50} and Gaussian  scheduled CIFAR-100 \citep{shanahan2021encoders}) demonstrate CLIP’s competitiveness on all these CL settings. This generalization behaviour is due to the large-scale pre-training of vision-language model like CLIP that optimizes contrastive training objective on 400M image-text pairs scraped from the internet. During pre-training, CLIP learns a diverse range of high-level representations which are transferable to multiple downstream tasks including the incremental tasks. We also show how simple prompt engineering for text inputs affects the CLIP's performance for CL.

In summary,  this work layouts the baseline for the future direction in continual learning based on the pre-trained vision-language models. We evaluate the pre-trained frozen CLIP model in a variety of continual learning settings on popular image recognition benchmarks and compare to current state-of-the-art methods to show that out-of-box CLIP representations perform competitively in all cases. Our results aim to consolidate the fragmented efforts in continual learning landscape that work on specific settings, highlighting the need for generic approaches that can work across multiple settings.

\section{Related Works}

\textbf{Continual Learning:} The existing continual learning methods mostly employ one of the following schemes: (1) model regularization (2) memory replay, and (3) dynamic network expansion. Model Regularization-based methods \citep{kirkpatrick2017overcoming, aljundi2018memory, kirkpatrick2017overcoming, li2017learning} avoid catastrophic forgetting by limiting the plasticity of the model parameters that are important for previous tasks. Even though these methods do not require memory replay, they have only been shown to work on simplistic task-incremental setting (the task identity is assumed to be known at inference time) and on smaller datasets \citep{wu2019large}. Memory Replay based methods use exemplars that are either stored in the memory, or synthesized using generative techniques, and are effective on more challenging settings and datasets \citep{rebuffi2017icarl,kamra2017deep, buzzega2020dark, cha2021co2l}. However, their performance degrades for smaller memory size \citep{cha2021co2l}, and storing these exemplars can introduce security and privacy concerns \cite{shokri2015privacy}. Architecture-driven CL methods either dynamically expand a network \cite{rusu2016progressive, li2019learn, zhao2022deep}, or divide into sub-networks to cater for the new tasks \citep{zhao2022deep, wang2020learn,ke2020continual, rajasegaran2019random}. Such approaches lack scalability, since the network capacity grows with tasks.

\textbf{Vision-language Models:} Training a joint vision-language embedding space enables interaction amongst text and image data, and is critical to solve problems such as zero-shot learning, visual grounding, and image captioning. While the initial vision-language models were single-stream and processed the concatenated input from visual and text data as a single set of input \citep{li2019visualbert, kim2021vilt}, more recent approaches, such as the Contrastive Language-Image Pre-training (CLIP) \citep{radford2021learning} are dual-stream with dedicated encoders for image and text inputs. The representations from the two encoders are projected into a unified embedding space, and a contrastive learning objective is employed to minimize the distance between matching image-caption pairs, and maximize otherwise. Subsequent works have shown that CLIP is scalable, and its capabilities improve once trained on large-scale noisy data of 1 billion non-curated samples \cite{jia2021scaling}. The representations learned by the CLIP have been shown to generalize well across numerous downstream tasks, including image-text retrieval, with excellent zero-shot transfer capabilities \citep{li2021align}.  CLIP can also be flexibly adapted to videos \citep{yuan2021florence, xu2021videoclip, fang2021clip2video}, and to capture object-level interactions \citep{yao2021filip, li2021align, Hanoona2022Bridging}. However, the applicability of CLIP representations for CL is not yet investigated. 

With recent advances in Vision Transformers \citep{khan2021transformers}, and prompt-based fine-tuning in NLP \citep{li2021prefix}, Wang \emph{et al.} have shown that interacting with an ImageNet pre-trained model via prompt learning is a promising approach for continual learning \citep{wang2022dualprompt, wang2022learning}. A small set of learnable parameters, called prompts, is appended to the input, and enables quick adaptation of a frozen ImageNet pre-trained model to new streaming tasks. In our analysis, we show that directly leveraging the pre-trained vision-language model, without introducing any learnable parameters, is a simple yet promising approach to continual learning. We argue that adapting a joint vision-language model like CLIP \citep{radford2021learning} for continual learning presents multiple advantages. It enables catering for practical scenarios where there are no well-defined task identities and boundaries, and the model is required to dynamically adapt to streaming data in a task-agnostic manner. As such, leveraging from the CLIP model requires no compute expensive training or fine-tuning on the data for new tasks. Further, in contrast to the current state-of-the-art methods that require a memory buffer to store training examples from previous tasks, Continual-CLIP approach is rehearsal free, and is more suitable for scenarios where storing training examples could have practical privacy and security concerns or storage constraints \citep{shokri2015privacy}. Instead of storing past samples in the memory buffer, where the performance can deteriorate for small buffer size, Continual-CLIP approach requires a constant memory throughout all the learning episodes. In summary, Continual-CLIP approach is memory free, does not require test-time task identity information, can be flexibly and easily adapted to any number of classes without requiring any additional learnable parameters.

\section{Methodology}

In this section, we first briefly discuss different continual learning settings. Next, we introduce Contrastive Language-Image Pre-training (CLIP, \cite{radford2021learning}) and explain how to apply it to different downstream continual learning tasks in a zero shot manner with hand-crafted prompts. 

\subsection{Continual Learning Formulation.}
\label{sec:background}

Different Continual Learning (CL) problems focus on training models on a non-stationary data from sequential tasks, while reducing forgetting on the old tasks\footnote{We use term task for a distinct training episode happening in CL.}. Consider a sequence of tasks $D = \{D_1, D_2, \dots, D_T\}$, where the $t^{th}$ task $D_{t} = \{(\mathbi{x}_{i}^{t}, y_{i}^{t})\}_{i = 1}^{N_{t}}$ contains tuples of input samples $\mathbi{x}_{i}^{t} \in \mathcal{X}$ and its corresponding label $y_{i}^{t} \in \mathcal{Y}$. The goal is to optimize the model $f_{\theta}: \mathcal{X} \rightarrow \mathcal{Y}$ parameterized by $\theta$, such that it predicts the label $y = f_{\theta}(\mathbi{x}) \in \mathcal{Y}$ given an unseen test sample $\mathbi{x}$ from an arbitrary task. During task $t$, the data from the previous distribution $D_{t - 1}$ is not available or restricted.

Based on the input $P(\mathcal{X}^{(t)})$ and output $P(\mathcal{Y}^{(t)})$ distributions of task $t$, with $P(\mathcal{X}^{(t)}) \neq P(\mathcal{X}^{(t+1)})$, continual learning can be classified into four popular settings with slightly different assumptions \emph{i.e.}, task-incremental, class-incremental, domain-incremental and task-free CL. The common task-, class-, domain-incremental settings assumes task data $D_{t}$ arrives in a sequence such that $t = \{1, 2, \dots, T\}$. At task $t$, the class-incremental setting defines output space for all observed class labels $\mathcal{Y}^{(t)} \subset \mathcal{Y}^{(t+1)}$ with $P(\mathcal{Y}^{(t)}) \neq P(\mathcal{Y}^{(t+1)})$. Different from class-incremental setting, task-incremental settings defines $\mathcal{Y}^{(t)} \neq \mathcal{Y}^{(t+1)}$ with $P(\mathcal{Y}^{(t)}) \neq P(\mathcal{Y}^{(t+1)})$ requires tasks label $t$ to indicate isolated output heads $\mathcal{Y}^{(t)}$. Different from task- and class-incremental settings where each task has different classes, domain incremental setting is defined as $\mathcal{Y}^{(t)} = \mathcal{Y}^{(t+1)}$ with $P(\mathcal{Y}^{(t)}) = P(\mathcal{Y}^{(t+1)})$ and $P(\mathcal{X}^{(t)}) \neq P(\mathcal{X}^{(t+1)})$ such that it contains a set of images drawn from a different domain, but has the same set of classes for every task. The more challenging setting is task-free (or task-agnostic) setting, where the task data $D_{t}$ changes smoothly and the task identity $t$ is unknown \citep{de2021continual}. In this work, we perform extensive experiments to show the effectiveness of CLIP based approach on the challenging class- and domain-incremental settings, as well as the task-agnostic setting in CL.

\subsection{Continual-CLIP}

Contrastive Language-Image Pre-training (CLIP, \cite{radford2021learning}) consists of two parallel encoders, one is for text and the other for images. The text encoder is based on the Transformer \citep{vaswani2017attention} architecture which generates embedding representations for the text-based language inputs. On the other hand, the image encoder architecture can be based on a CNN \emph{e.g.}, ResNet-50 \cite{he2016deep} or a Vision Transformer (ViT) model \citep{dosovitskiy2020image} to transforms the high-dimensional input images into a compact embedding space. The embedding feature dimension for the text and image encoder are same, thus enabling learning a shared and unified representation space for the two  modalities. 

The CLIP model is trained with a contrastive loss which promotes the similarities between image and text embeddings belonging to the same image-caption pair, so that both get aligned in the joint feature space. Given a batch of image-text pairs, CLIP objective is to maximize the cosine similarity between matched pairs while minimize the similarity between unmatched image-text embedding pairs. Using this learning objective, the model is trained on a large scale dataset of 400M image-caption pairs, and it learns highly transferable representations for image and text data, which have demonstrated impressive zero-shot generalization capabilities.

Let us denote CLIP model as $\mathcal{F} = \{\mathbf{E}_\texttt{visual}, \mathbf{E}_\texttt{text}\}$, where $\mathbf{E}_\texttt{visual}$ and $\mathbf{E}_\texttt{text}$ are image and text encoders respectively. Consider a  $K$-class classification problem, such that a single test image $\mathbi{x}_{test} \in \mathbb{R}^{C \times H \times W}$ belongs to a class $y \in \mathbb{R}^{K}$.  In the traditional zero-shot classification scenario, every $y_i \in \mathcal{Y} = \{y_1, y_2, \dots, y_K\}$ is prepended by a hand-crafted prompt template $\mathbi{p}$, such as $~``\txt{a photo of a \{category\}}''$ to form a category-specific text input $\{\mathbi{p}; y_{i}\}$. Then this text input is fed to the text-encoder to get the text embedding $\{\mathbi{t}_1, \mathbi{t}_2, \dots, \mathbi{t}_K\}$, where $\mathbi{t}_i = \mathbf{E}_\txt{text}(\{\mathbi{p}; y_{i}\})$. The image input $\mathbi{x}_{test}$ is fed to the $\mathbf{E}_\txt{visual}$ to get the corresponding image embedding $\mathbi{v} = \mathbf{E}_\txt{visual}(\mathbi{x}_{test})$. Both the text and image embeddings are then matched to compute the similarity score $\mathbi{s}_i = \txt{sim}(\mathbi{t}_i\cdot\mathbi{v})$, where $\txt{sim}(\cdot,\cdot)$ denotes the cosine similarity. The prediction probability of $\mathbi{x}_{test}$ can be denoted by $p(y_{i} | \mathbi{x}_{test}) = \frac{\txt{exp}(\txt{sim}(\mathbi{t}_i\cdot\mathbi{v}))}{\sum_{i=1}^{K}\txt{exp}(\txt{sim}(\mathbi{t}_i\cdot\mathbi{v}))}$.

The traditional classifiers learn closed-set visual concepts, whereas the vision-language pre-trained models like CLIP learn open-set visual concepts via the high-capacity text encoder. This leads to a broader semantic space, making learned representation more transferable to downstream tasks. The pre-trained vision-language CLIP model therefore offers impressive zero-shot capabilities. We leverage the frozen CLIP model for image recognition in continual learning settings without any training in a strictly zero-shot fashion. For evaluation, we first initialize both the CLIP encoders and freeze the weight. Different from traditional evaluation where all the test dataset will be fed to the model, in continual evaluation we assume that the model is learning a current task $t$ and evaluated only on the tasks observed so far.

\section{Experiments}

\subsection{Experimental Protocols}

\textbf{Datasets:} We evaluate Continual-CLIP on seven different datasets and 13 different learning task configurations. For class-incremental settings, we evaluate Continual-CLIP on CIFAR-100, ImageNet-100 \& 1K, TinyImageNet under different class splits. \textbf{(a)} In CIFAR-100, we compare the performance on 10 steps (10 new classes per step), 20 steps (5 new classes per step), and 50 steps (2 new classes per step) \citep{Douillard_2022_CVPR, Yan_2021_CVPR}. \textbf{(b)} In ImageNet-100, we consider two evaluation settings; \textit{ImageNet-100-B0} which has the same number of classes for all the steps (i.e., 10 classes per step) and \textit{ImageNet-100-B50} that contains 50 classes for the first step and the rest of the 50 classes are observed incrementally in the next 10 steps (5 classes per steps) \citep{Yan_2021_CVPR}. \textbf{(c)} We divided ImageNet-1K into 10 incremental steps (100 classes per step) \citep{Yan_2021_CVPR}. \textbf{(d)} In TinyImageNet, we kept 100 classes for the first (or base) step, and then similar to ImageNet-100, add the rest of the classes incrementally in three different number of steps i.e., 5 steps (20 classes per step), 10 steps (10 classes per step), and 20 steps (5 classes per step) \citep{zhu2021prototype}. Note that we do not evaluate under task-incremental setting where task IDs are known at inference since it is the easiest of the continual learning settings.

We evaluate Continual-CLIP in domain-incremental setting on CORe50 and CLEAR-10 \& 100 datasets. The CORe50 has 11 different scenarios and 10 classes. Out of which we exclude 8 domains since we are not training (or fine-tuning) the model and only consider 3 domains for evaluation following the guidelines in \citep{lomonaco2017core50}. For the experiment on CLEAR-10 \& 100, we use the Avalanche \citep{lomonaco2021avalanche} pre-build methods to get the evaluation scenarios. 

We used the Gaussian schedule CIFAR-100 for the evaluation in task-free (or task-agnostic) setting. We use the same setup as \citep{shanahan2021encoders} to check the effectiveness of Continual-CLIP method in task-agnostic setting. For the evaluation, \citep{shanahan2021encoders} keep the evaluation dataset aside and do evaluation on the whole dataset after every task. Since Continual-CLIP model is frozen, we report the final test accuracy for comparison.

\textbf{Evaluation Metrics:} In class-incremental setting, we compare Continual-CLIP with other baseline approaches in terms of ``Avg" accuracy, which is the average of all accuracy values obtained at each time step and ``Last" accuracy which is the final accuracy after learning all the tasks. In domain-incremental setting, we used ``In-domain",``Next-domain", ``Forward transfer", ``Backward transfer", and ``Overall accuracy" to compare Continual-CLIP. Note that for the CORe50 dataset, only a single test set is available, so we report the accuracy on this set. Similarly, for the task-agnostic setting, we only report test accuracy.

\textbf{Implementation Details:} We use the official CLIP \citep{radford2021learning} implementation in zero-shot evaluation settings. To build continual scenarios for class-incremental setting, we heavily used Continuum \citep{douillardlesort2021continuum} and follow the same evaluation setting from \cite{Douillard_2022_CVPR}. For the domain-incremental scenarios on CORe50 and CLEAR datasets, we use the Avalanche library \citep{lomonaco2021avalanche}. Since the CIFAR100 and TinyImageNet datasets contain low resolution images, we use a ``\txt{a bad photo of a \{category\}}" prompt template. The datasets like ImageNet-100 and 1K have high resolution images so we  use ``\txt{a good photo of a \{category\}}" for the evaluation of Continual-CLIP for all the experiments.

For the evaluation, we used the same data-reprocessing technique as defined in \citep{radford2021learning}, which include Bicubic interpolation, Center Cropping, and Normalization.

\subsection{Results}

\begin{table}[t]
  \begin{center}
    \caption{Comparison of state-of-the-art CL methods on CIFAR100 benchmark in class-incremental setting, in terms of the average and last task accuracy values.}
    \label{tab:cifar100_class_incre}
    {\small
    \begin{tabular}{{l}*{5}{c}{c}} 
    \toprule
        \rowcolor{lightblue!10} & \multicolumn{2}{c}{{\cellcolor{lightblue!50} 10 steps}} & \multicolumn{2}{c}{{\cellcolor{lightblue!10} 20 steps}} & \multicolumn{2}{c}{\cellcolor{lightblue!50} 50 steps}  \\
    \rowcolor{lightblue!10} \textbf{Methods} & \textbf{Avg} & \textbf{Last} & \textbf{Avg} & \textbf{Last} & \textbf{Avg} & \textbf{Last} \\
    \midrule
    iCaRL \citep{rebuffi2017icarl}           & 65.27 & 50.74 & 61.20 & 43.74 & 56.08 & 36.62 \\
    UCIR \citep{hou2019learning}             & 58.66 & 43.39 & 58.17 & 40.63 & 56.86 & 37.09 \\
    BiC \citep{wu2019large}                  & 68.80 & 53.54 & 66.48 & 47.02 & 62.09 & 41.04 \\
    RPSNet  \citep{rajasegaran2019adaptive}  & 68.60 & 57.05 & -     & -     & -     & -     \\
    WA \citep{zhao2020maintaining}          & 69.46 & 53.78 & 67.33 & 47.31 & 64.32 & 42.14 \\
    PODNet \citep{douillard2020podnet}       & 58.03 & 41.05 & 53.97 & 35.02 & 51.19 & 32.99 \\
    DER (w/o P) \citep{Yan_2021_CVPR}        & \textbf{75.36} & 65.22 & 74.09 & 62.48 & 72.41 & 59.08 \\
    DER \citep{Yan_2021_CVPR}                & 74.64 & 64.35 & 73.98 & 62.55 & 72.05 & 59.76 \\
    DyTox \citep{Douillard_2022_CVPR}        & 67.33 & 51.68 & 67.30 & 48.45 & 64.39 & 43.47 \\
    DyTox+ \citep{Douillard_2022_CVPR}       & 74.10 & 62.34 & 71.62 & 57.43 & 68.90 & 51.09 \\
    \midrule
    \rowcolor{purple!5} Continual-CLIP               & 75.17 & \textbf{66.72} & \textbf{75.95} & \textbf{66.72} & \textbf{76.49} & \textbf{66.72} \\
    \bottomrule
    \end{tabular}}
\end{center}
\end{table}

\begin{figure}
  \includegraphics[width=\linewidth]{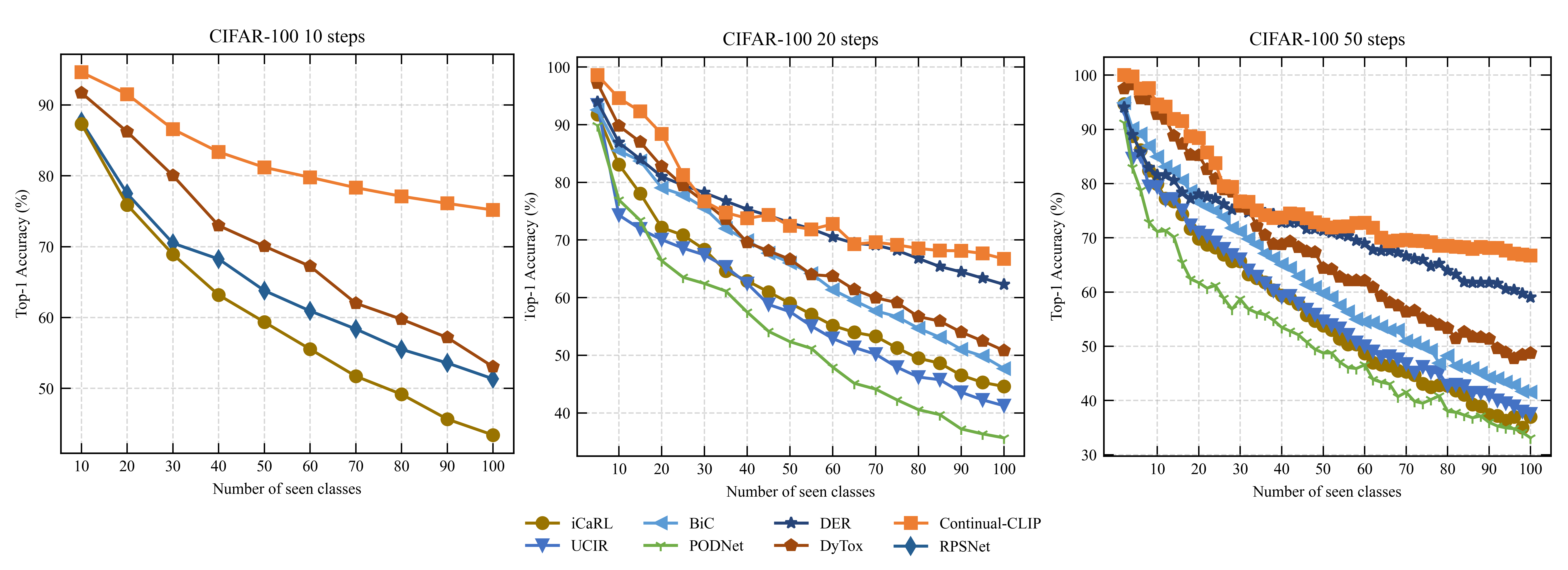}
  \caption{Step-wise performance trends on the CIFAR-100 for 10, 20, and 50 steps. Continual-CLIP performs well especially for longer incremental episodes. Note that the close competitor DER expands the architecture with new tasks, thereby significantly increasing compute complexity.}
  \label{fig:cifar100_acc_plot}
\end{figure}

\textbf{Class-incremental Setting.} The results in Table \ref{tab:cifar100_class_incre} compare Continual- with other baselines on CIFAR-100 dataset in three different settings (10, 20 and 50 steps). In 10 steps setting, even without any training or fine-tuning, Continual-CLIP achieves competitive results in terms average and last accuracy, compared with the recent state-of-the-art methods such as DyTox \citep{Douillard_2022_CVPR} and DER \citep{Yan_2021_CVPR}. Specifically, In 20 steps setting, Continual-CLIP reaches 75.95\% in ``Avg" accuracy, and for the 50 steps setting, it reaches 76.49\% in ``Avg" accuracy. The last accuracy is same for all the cases (since it is zero-shot evaluation). The results suggest that Continual-CLIP consistently performs better than majority of the compared methods by a significant margin. This trend is especially notable on large number of tasks, which is a harder case in CL (Figure~\ref{fig:cifar100_acc_plot}).

\begin{table}[t]
  \begin{center}
    \caption{Comparison of state-of-the-art CL methods on different ImageNet benchmarks, in class-incremental settings with 10 splits, in terms of average and last accuracy values.}
    \label{tab:imagenet_class_incre}
    {\small
    \begin{tabular}{{l}*{5}{c}{c}} 
        \toprule
        \cellcolor{lightblue!10}   & \multicolumn{2}{c}{{\cellcolor{lightblue!50} ImageNet100-B0}} & \multicolumn{2}{c}{\cellcolor{lightblue!10}ImageNet1K}    & \multicolumn{2}{c}{\cellcolor{lightblue!50}ImageNet100-B50}  \\
        \rowcolor{lightblue!10} \textbf{Methods}    & \textbf{Avg}   & \textbf{Last}    & \textbf{Avg}   & \textbf{Last}    & \textbf{Avg}   & \textbf{Last}  \\
        \midrule
        iCaRL \citep{rebuffi2017icarl}              & -     & -     & 38.40 & 22.70 & -     & -     \\
        UCIR \citep{hou2019learning}                & -     & -     & -     & -     & 68.09 & 57.30 \\
        WA \citep{zhao2020maintaining}              & -     & -     & 65.67 & 55.60 & -     & -     \\
        TPCIL \citep{tao2020topology}               & -     & -     & -     & -     & 74.81 & 66.91 \\
        PODNet \citep{douillard2020podnet}          & -     & -     & -     & -     & 74.33 & -     \\
        Simple-DER \citep{Zhuoyun2021preserve}      & -     & -     & 66.63 & 59.24 & -     & -     \\
        DER (w/o P) \citep{Yan_2021_CVPR}           & 77.18 & 66.70 & 68.84 & 60.16 & 78.20 & 74.92 \\
        DER \citep{Yan_2021_CVPR}                   & 76.12 & 66.06 & 66.73 & 58.62 & 77.13 & 72.06 \\
        DyTox \citep{Douillard_2022_CVPR}           & 73.96 & 62.20 & -     & -     & -     & -     \\
        DyTox+ \citep{Douillard_2022_CVPR}          & 77.15 & 67.70 & 70.88 & 60.00 & -     & -     \\
        \midrule        
        \rowcolor{purple!5} Continual-CLIP   & \textbf{85.00} & \textbf{75.42} & \textbf{75.51} & \textbf{67.71} & \textbf{79.69} & \textbf{75.42} \\
        \bottomrule
    \end{tabular}}
\end{center}
\end{table}
    
The results on the ImageNet-100 \& 1K datasets are presented in Table~\ref{tab:imagenet_class_incre}. In standard settings (i.e., same number of classes in all steps; see columns 2-5), Continual-CLIP shows an improvement of 7.84\% on ImageNet-100 and 4.63\% on ImageNet-1K dataset in terms of the ``Avg" accuracy compared with the second best method. Similarly, a significant improvement of 7.71\% on ImageNet-100 and 7.70\% on ImageNet-1K  in ``Last" accuracy is achieved. For the other setting with 50 base classes (i.e., ImageNet-100-B50), Continual-CLIP shows better results than the recent methods.

\begin{table}
  \begin{center}
    \caption{Comparison of state-of-the-art CL methods on different TinyImageNet splits in class-incremental settings with 50 base classes, in terms of the average and last accuracy values.}
    \label{tab:tiny_imagenet_class_incre}
    {\small
    \begin{tabular}{{l}*{5}{r}{r}} 
   \toprule
    \cellcolor{lightblue!10}  & \multicolumn{2}{c}{{\cellcolor{lightblue!50} 5 steps}} &  \multicolumn{2}{c}{\cellcolor{lightblue!10} 10 steps}    &  \multicolumn{2}{c}{\cellcolor{lightblue!50} 20 steps}  \\
    \rowcolor{lightblue!10} \textbf{Methods}    & \textbf{Avg}   & \textbf{Last}    & \textbf{Avg}   & \textbf{Last}    & \textbf{Avg}   & \textbf{Last}  \\
    \midrule
    EWC \citep{kirkpatrick2017overcoming}       & 19.01 & 6.00  & 15.82 & 3.79  & 12.35 & 4.73 \\
    LwF \citep{li2017learning}                  & 22.31 & 7.34  & 17.34 & 4.73  & 12.48 & 4.26 \\
    LwF-MC \citep{rebuffi2017icarl}             & 29.09 & 15.63 & 23.03 & 13.25 & 17.31 & 7.95 \\
    iCaRL \citep{rebuffi2017icarl}              & 34.27 & 23.22 & 30.94 & 20.82 & 27.83 & 20.16 \\
    iCaRL-NCM \citep{rebuffi2017icarl}          & 45.95 & 34.60  & 43.22 & 33.22 & 37.85 & 27.54 \\
    EEIL \citep{castro2018end}                  & 47.17 & 35.12 & 45.03 & 34.64 & 40.41 & 29.72 \\
    UCIR \citep{hou2019learning}                & 50.30 & 39.42 & 48.58 & 37.29 & 42.84 & 30.85 \\
    MUC \citep{liu2020more}                     & 32.23 & 19.20  & 26.67 & 15.33 & 21.89 & 10.32 \\
    PASS \citep{zhu2021prototype}               & 49.54 & 41.64 & 47.19 & 39.27 & 42.01 & 32.93 \\
    DyTox \citep{Douillard_2022_CVPR}           & 55.58 & 47.23 & 52.26 & 42.79 & 46.18 & 36.21 \\
    \midrule
    \rowcolor{purple!5} Continual-CLIP  & \textbf{70.49} & \textbf{66.43} & \textbf{70.55} & \textbf{66.43} & \textbf{70.51} & \textbf{66.43} \\
    \bottomrule
    \end{tabular}}
\end{center}
\end{table}

\begin{figure}
  \includegraphics[width=\linewidth]{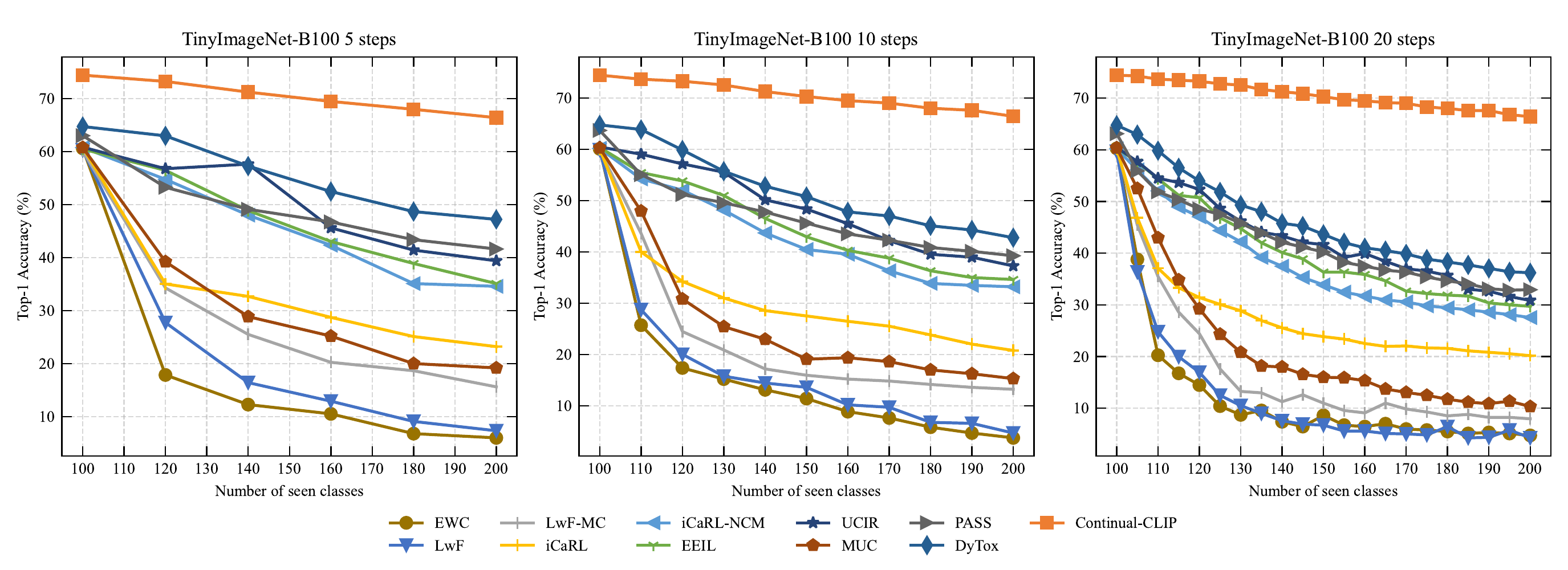}
  \caption{Accuracy trends on the TinyImageNet dataset in three different settings of step sizes. Note that the other competing methods require training at each stage, use memory buffers, may not apply to all CL settings and/or dynamically expand the architecture to learn new tasks. }
  \label{fig:tiny_imagenet_acc}
\end{figure}

From the results reported in Table~\ref{tab:tiny_imagenet_class_incre}, we observe that on TinyImageNet dataset with base class of 100, Continual-CLIP shows consistent improvements  in all 3 settings. Continual-CLIP shows on average 20.30\% improvement compared with the second best method DyTox in terms of the average accuracy, also it shows impressive gains in terms of the last accuracy.

\begin{table}[t]
  \begin{center}
    \caption{Domain incremental setting comparison with winning team of the recent CVPR 2022 Continual LEArning on Real-World Imagery (CLEAR) Challenge \citep{aicrowd2022clear, lin2021clear}.}
    \label{tab:domain_incremental_clear}
      {\small
    \begin{tabular}{{l}*{5}{r}{r}} 
   \toprule
    \rowcolor{lightblue!10} \textbf{Datasets}    & \textbf{Methods}   & \multicolumn{1}{p{1cm}}{\textbf{Overall Acc}}   & \multicolumn{1}{p{1cm}}{\textbf{Next-domain}}   & \multicolumn{1}{p{1cm}}{\textbf{In-domain}}   & \textbf{Backward}   & \textbf{Forward}  \\
    \midrule
    \multirow{2}{*}{CLEAR-10}       & Top-1 team    & 92.70  & 92.50 & 93.40 & \textbf{94.20} & 90.90 \\
                                    & Continual-CLIP & \textbf{93.79}  & \textbf{93.51} & \textbf{93.58} & 93.83 & \textbf{93.33} \\
    \midrule
    \multirow{2}{*}{CLEAR-100}      & Top-1 team & 91.46 & 91.25 & 91.99 & 93.40 & 89.20 \\
                                    & Continual-CLIP & \textbf{93.63} & \textbf{93.50} & \textbf{93.50} & \textbf{93.66} & \textbf{93.33} \\
    \bottomrule
    \end{tabular}}
\end{center}
\end{table}

\begin{table}
  \begin{center}
    \caption{Core50 dataset comparisons with other baselines in domain incremental setting. The values for compared methods are from \citep{lomonaco2017core50, parisi2018lifelong}.}
    \label{tab:domain_incremental_core50}
    {\small
    \begin{tabular}{{l}{r}} 
    \toprule
    \rowcolor{lightblue!10} \textbf{Methods}    & \textbf{Test Acc (\%)}  \\
    \midrule
    Na\"ive \citep{lomonaco2017core50}          & 54.69 \\
    EWC \citep{kirkpatrick2017overcoming}       & 57.40 \\
    LwF \citep{zhao2020maintaining}             & 59.42 \\
    Cumulative \citep{lomonaco2017core50}       & 65.15 \\
    GDM \citep{parisi2018lifelong}               & 74.87 \\
    \midrule
    \rowcolor{purple!5} Continual-CLIP   & \textbf{84.73} \\
    
    \bottomrule
    \end{tabular}}
\end{center}
\end{table}

\textbf{Domain-incremental setting.} Table \ref{tab:domain_incremental_clear} shows the comparison of Continual-CLIP method with the winning entry of the CVPR 2022 CLEAR Challenge Leaderboard \citep{aicrowd2022clear, lin2021clear}. The values are reported on the test set following the evaluation protocol in \citep{lomonaco2017core50}. For CLEAR-10 \& 100, Continual-CLIP performs competitively with the top performing team.

\textbf{Task-agnostic setting.} This is a general setting in which there is no constraint of task, class, or domain and is considered to be quite challenging. This setting is relatively under-explored in the literature. In Table~\ref{tab:task_agnostic_cifar100}, we show a comparison of Continual-CLIP with the previous best method i.e., Encoders and Ensemble \citep{shanahan2021encoders}. Continual-CLIP outperforms the compared approach with a large margin without any need of training or model ensembles.

\begin{table}
  \begin{center}
    \caption{Comparisons in Task-agnostic setting on Core50 dataset. 
    }
    \label{tab:task_agnostic_cifar100}
    {\small
    \begin{tabular}{{l}{r}} 
    \toprule
    \rowcolor{lightblue!10} \textbf{Methods}            & \textbf{Test Acc (\%)}  \\
    \midrule
    Encoders and Ensemble \citep{shanahan2021encoders}  & 39.0 \\
    \rowcolor{purple!5} Continual-CLIP                  & \textbf{66.72} \\
    
    \bottomrule
    \end{tabular}}
\end{center}
\end{table}

In summary, our extensive empirical evaluations and comparisons provide an evidence that the  Continual-CLIP consistently shows impressive results in all continual learning settings, without the need of any fine-tuning (or training), dedicated memory for past exemplars and model copies, complex hyper-parameter tuning, dynamic model expansion or changing classification heads. 

 \subsection{Analysis on Text Prompts}
 
  \begin{table}[t]
  \begin{center}
    \caption{Effect of using different class names on the Continual-CLIP accuracy on ImageNet-1k with a prompt ``\texttt{a photo of a \{\}.}"}
    \label{tab:different_class_names}
    {\small
    \begin{tabular}{{l}{r}{r}} 
    \toprule
    \rowcolor{lightblue!10} \textbf{Class Names Type}    & \textbf{Avg Acc (\%)} & \textbf{Last Acc (\%)}  \\
    \midrule
    (\texttt{a}) ImageNet default                           & 72.96 & 64.44 \\
    (\texttt{b}) \cite{radford2021learning} curated         & 74.81 & 66.58 \\
    (\texttt{c}) First synonym from each subset             & 71.65 & 63.97 \\
    
    \bottomrule
    \end{tabular}}
\end{center}
\end{table}

 \begin{table}[t]
  \begin{center}
    \caption{The performance of Continual-CLIP on 80 different prompts with \txt{(p1) decision-based pooling} (the average of all 80 prompts) and \txt{(p2) embedding pooling} on ImageNet-100 dataset.}
    \label{tab:different_prompts}
    {\small
    \begin{tabular}{{l}{r}{r}} 
    \toprule
    \rowcolor{lightblue!10} \textbf{Class Names Type}    & \textbf{Acc (\%)} & \textbf{Last Acc (\%)}  \\
    \midrule

    Decision-based pooling    & 82.24 & 72.11 \\
    Decision-based pooling (top-10)   & 84.63 & 74.94 \\
    Embedding pooling         & 84.85 & 75.46 \\
    
    \bottomrule
    \end{tabular}}  

\end{center}
\end{table}

 In this section, we investigate the effect of different class names used as part of text prompts on the Continual-CLIP accuracy's. We found different types of naming convention for the ImageNet-1K dataset classes, and use them to evaluate Continual-CLIP performance. Specifically, we used three names from three different sources: (\textbf{a}) ImageNet original labels, where each of the 1000 ImageNet \citep{deng2009imagenet} classes correspond to a WordNet synset (a set of synonyms), (\textbf{b}) the original ImageNet labels have overlapping meaning (e.g., ``\txt{nail}", CLIP model understood it has ``\txt{fingernail}" so it was changed to ``\txt{metal nail}") so  \cite{radford2021learning} curated the default labels to overcome this confusion, and (\textbf{c}) the labels with the first synonym from each Synset. Table \ref{tab:different_class_names} shows the effect of different class names conventions on model's performance. With class name type (\textbf{b}) model achieves better results than the other two class types, this is because the class names for (\textbf{b}) is curated in such a way that there will be a clear boundary and distinction between different class names, and for other two types there can be overlapping class meanings.
 
 We further explore the effectiveness of Continual-CLIP with the prompt engineering. For that, we used different prompts and analyse the results on ImageNet-100 dataset. In Table~\ref{tab:different_prompts}, we report the results for two different prompt techniques for continual zero-shot CLIP. \textbf{(p1) Decision-based pooling} - We compute the score for each prompt separately and then average pool them. \textbf{(p2) Embedding pooling} - We first compute the text-embedding for each prompt by passing it through the text-encoder and then create a single classifier head by stacking all the embeddings. The results in Table \ref{tab:different_prompts} show the trade-off between embedding and decision-based pooling. Embedding and Decision-based pooling needs to collect several prompt templates and requires additional computation to get the text-embedding for each prompt as compared to Continual-CLIP main results. Also, with the decision and embedding pooling, we need domain expert knowledge for prompt-engineering, and repeat the process over multiple trials until we can find best performing prompt for the given scenario.

\begin{figure}
    \hfill
    \subfigure{
    \includegraphics[width=5cm]{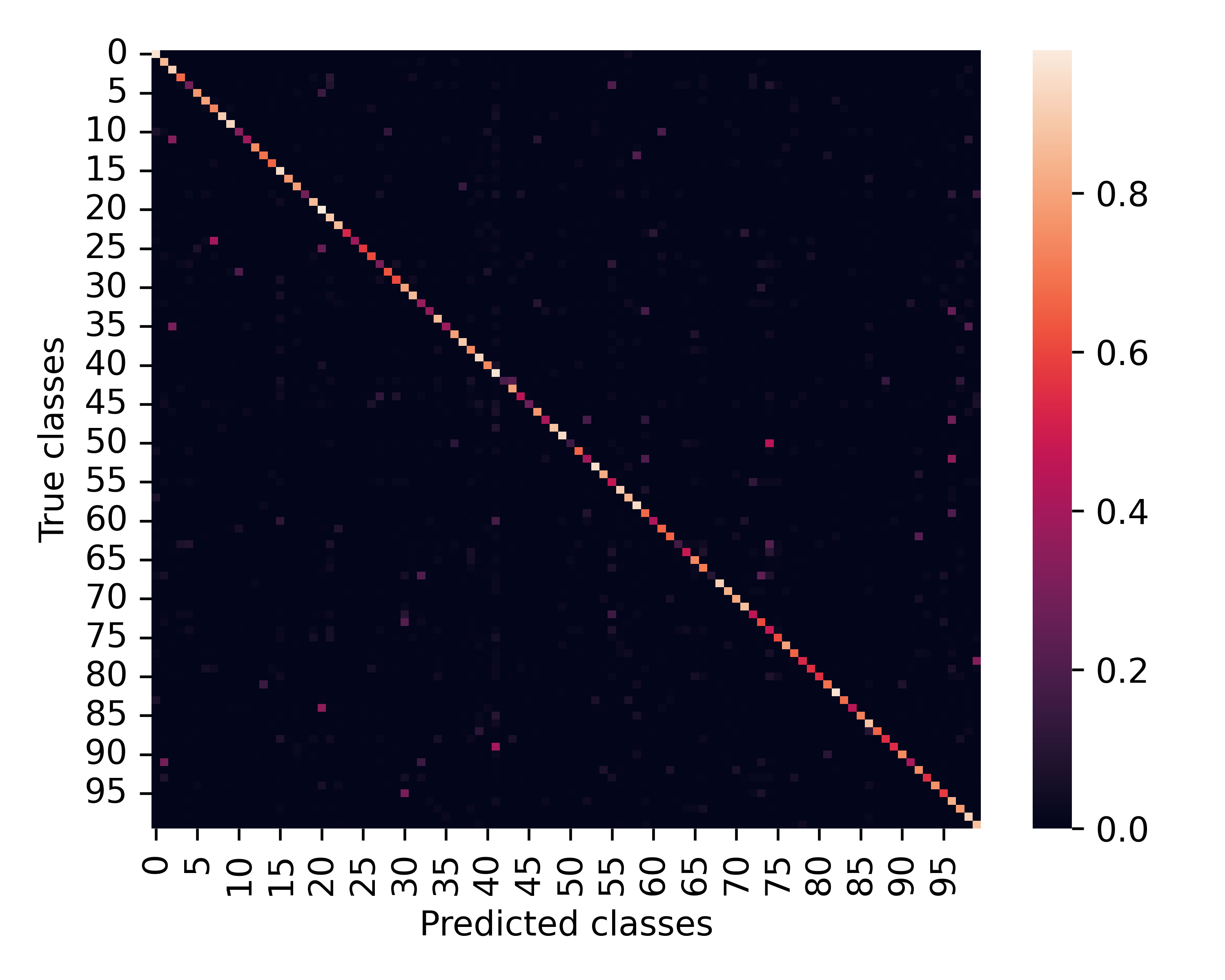}
    }
    \hfill
    \subfigure{\includegraphics[width=5cm]{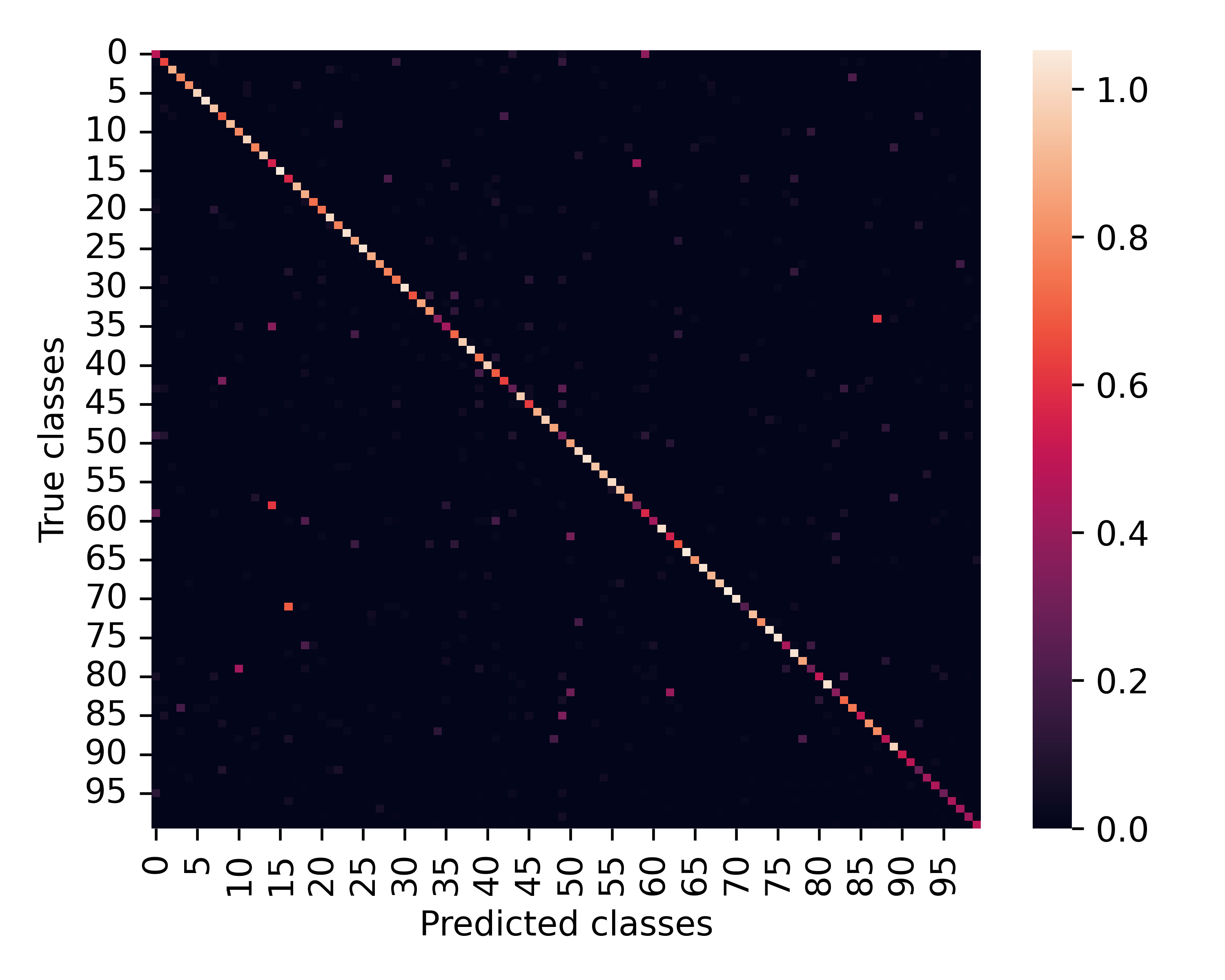}}
    \hfill
    \caption{\textit{Left:} Confusion matrix for CIFAR-100 dataset. \textit{Right:} Confusion matrix for ImageNet-100 dataset. We note that majority errors occur between semantically similar classes. }
    \label{fig:confusion_matrix}
\end{figure}

\section{Conclusion}

In this work, we evaluate a simple yet effective baseline that consistently achieves favorable results for three different incremental learning settings; (1) class-incremental, (2) domain-incremental, and (3) task-agnostic incremental learning. Continual-CLIP is a standard CLIP model evaluated to work in  continual learning scenarios. The experiments on different benchmarks with challenging configurations shows that the Continual-CLIP outperforms the current state-of-the-art methods in continual learning without the need of any fine-tuning, replay buffers, or memory overhead. Continual-CLIP model can be used in any continual learning setting with zero or little modification. The state-of-the-art performance on the large datasets like ImageNet-1k (in class-incremental), CORe50 (in domain-incremental) shows that Continual-CLIP is scalable.     

Although Continual-CLIP shows excellent performance in the continual learning settings, there is a possibility of potential information leakage, \emph{e.g.}, CLIP might have already come across some of the classes from the evaluated downstream dataset during the pre-training phase. Also, there arise new questions based on their behaviour, \emph{e.g.}, it can be seen from the confusion matrix in Figure \ref{fig:confusion_matrix} (left: CIFAR100) the model gets confused when there is a close semantic resemblance: the class name for the class index ``\txt{50}" is ``\txt{mouse}" and the model is predicts it as ``\txt{74}" which is ``\txt{shrew}".

In future work, this simple baseline approach built on top of zero-shot transfer capabilities of CLIP can be extended with fast adaptation methodologies for downstream continual tasks. Current work provides grounds for the future development in the continual learning leveraging the vision-language foundational models. Our work also motivates rethinking the progress made so far in the continual learning problems where the state of the art methods come with several constraints and promotes looking for generic solutions that transcend beyond narrow settings and cumbersome memory and compute requirements. 

\bibliography{arxiv_preprint}
\bibliographystyle{arxiv_preprint}

\end{document}